\documentclass{article}
\usepackage{ijcai25}

\usepackage{times}
\usepackage{soul}
\usepackage{url}
\usepackage[hidelinks]{hyperref}
\usepackage[utf8]{inputenc}
\usepackage[small]{caption}
\usepackage{graphicx}
\usepackage{amsmath}
\usepackage{amsfonts}
\usepackage{amsthm}
\usepackage{amssymb}
\usepackage{booktabs}
\usepackage{algorithm}
\usepackage{algorithmic}
\usepackage[switch]{lineno}
\usepackage{pifont}   % For special symbols
\usepackage{amsthm}
\usepackage{booktabs} % 三线表 
\usepackage{multirow} % 合并行  
\usepackage{lipsum} % For dummy text
\usepackage{wrapfig} % For text wrapping around figures/tables

\usepackage{adjustbox}
\usepackage{booktabs} % For better-looking tables
\usepackage{siunitx}  % 用于格式化数字和单位  
\usepackage{wrapfig}
\usepackage{enumitem} % 加载 enumitem 包
\title{PeSANet: Physics-encoded Spectral Attention Network for\\ Simulating PDE-Governed Complex Systems}

\author{
Han Wan$^1$\thanks{Han Wan and Rui Zhang contributed equally to this work.}\and
Rui Zhang$^1$\and
Qi Wang$^1$\And
Yang Liu$^2$\And
Hao Sun$^1$\thanks{Corresponding author}
\\
\affiliations
$^1$Gaoling School of Artificial Intelligence, Renmin University of China, Beijing, China\\
$^2$School of Engineering Science, University of Chinese Academy of Sciences, Beijing, China\\
\emails
\{wanhan2001, rayzhang, qi\_wang, haosun\}@ruc.edu.cn,
liuyang22@ucas.ac.cn
}

\begin{document}

\maketitle

\begin{abstract}
Accurately modeling and forecasting complex systems governed by partial differential equations (PDEs) is crucial in various scientific and engineering domains. However, traditional numerical methods struggle in real-world scenarios due to incomplete or unknown physical laws. Meanwhile, machine learning approaches often fail to generalize effectively when faced with scarce observational data and the challenge of capturing local and global features. To this end, we propose the Physics-encoded Spectral Attention Network (PeSANet), which integrates local and global information to forecast complex systems with limited data and incomplete physical priors. The model consists of two key components: a physics-encoded block that uses hard constraints to approximate local differential operators from limited data, and a spectral-enhanced block that captures long-range global dependencies in the frequency domain. Specifically, we introduce a novel spectral attention mechanism to model inter-spectrum relationships and learn long-range spatial features. Experimental results demonstrate that PeSANet outperforms existing methods across all metrics, particularly in long-term forecasting accuracy, providing a promising solution for simulating complex systems with limited data and incomplete physics.
\end{abstract}

\section{Introduction}

Accurately modeling and simulating complex systems is crucial in various scientific and engineering fields, such as fluid dynamics, meteorology, and biology \cite{anderson1995computational,holton2013introduction,ghergu2011nonlinear}. These systems are typically described by partial differential equations (PDEs) that govern their spatiotemporal evolution. However, forecasting these systems efficiently and accurately in real-world scenarios presents significant challenges. Firstly, the physical laws that govern these systems can be incomplete or unknown, making it difficult to formulate the correct PDEs. Secondly, observational data are often scarce, limiting the ability to learn the underlying physics. Additionally, the characteristics of complex systems are multi-faceted, with local variants and global dependencies that need to be captured for reliable long-term predictions.

Over the years, various methods have been developed to address these challenges. Traditional numerical methods \cite{darwish2016finite,karniadakis2005spectral,zienkiewicz2005finite} enjoy good convergence and stability when PDEs are well-defined, but become difficult to apply when the system's governing laws are incomplete. To overcome this, data-driven methods such as DeepONet \cite{lu2021learning} and Fourier Neural Operators (FNOs) \cite{li2020fourier} use neural networks as surrogate models, aiming to learn the underlying physical evolution from data. These methods do not require explicit knowledge of the physical laws and achieve significant speedup compared to traditional numerical solvers. However, they often fail to offer meaningful predictions when the training data is limited, which is a common issue in real-world scenarios.

To address the limitations of purely data-driven approaches, researchers have developed a series of physics-aware methods that incorporate prior physical knowledge into neural networks. These methods can be classified into physics-informed and physics-encoded, depending on how they integrate the physical priors. Physics-informed methods typically use soft constraints, such as PDE residual-based loss functions, to enforce the model that aligns with the governing PDEs. Representative methods include Physics-Informed Neural Networks (PINNs) \cite{raissi2019physics} and Physics-Informed Neural Operators (PINOs) \cite{li2024physics}. However, due to their soft-constrained nature, these methods encounter challenges in hyperparameter tuning and training stability, and require sufficient prior knowledge of the governing PDEs \cite{krishnapriyan2021characterizing}. To this end, physics-encoded methods use a hard-constrained strategy to incorporate physical prior into the model architecture. For instance, the Physics-encoded Recurrent Convolutional Neural Networks (PeRCNN) \cite{Rao_2023,rao2022discovering} use specially designed convolutional kernels to approximate finite difference stencils. These approaches can automatically learn local differential operators that match the observed data, even when the governing PDE is unknown. However, due to their limited receptive field, they can only capture local variations and struggle to learn long-range spatial dependencies in complex systems.

To overcome these limitations, we propose the Physics-encoded Spectral Attention Network (PeSANet). The model consists of the physics-encoded and spectral-enhanced blocks, which learn the local and global evolution of the complex system, respectively. The physics-encoded block employs hard constraints to approximate local differential operators. This hard-constrained approach does not require sufficient physical prior knowledge and can achieve generalization from the limited observed data. On the other hand, the spectral-enhanced block handles spectral information in the frequency domain to capture global features. In particular, we proposed the spectral attention mechanism to exploit the inter-spectrum relationship, further enhancing the ability to learn long-range spatial dependencies. In summary, we make the following contributions:
\begin{itemize}
    \item We propose the Physics-encoded Spectral Attention Network (PeSANet) to integrate both local and global learning mechanisms, which can achieve long-term forecasting for complex systems with limited data and incomplete physical priors.
    \item We propose a spectral attention mechanism in the frequency domain to capture long-range spatial features. By using all frequency components and capturing interactions across spectral layers, this approach can enhance the model's ability to describe global system features.
    \item PeSANet achieves superior long-term prediction performance across multiple complex systems. Furthermore, it outperforms other baseline models in all metrics, showing robust generalization with limited training data and transfer learning ability across varying conditions.
\end{itemize}

\section{Related Work}

The numerical solution of PDEs has been a central focus of scientific computing for decades. Traditional numerical methods, such as finite difference and finite element methods, have achieved remarkable success in many applications. However, these methods can be computationally expensive and require significant domain knowledge. In recent years, deep learning has emerged as a powerful tool for solving PDEs, offering the potential for more efficient and flexible solutions. Existing AI-driven methods can be categorized into data-driven, physics-aware, and hybrid models. The key distinction lies in the incorporation of prior physical knowledge.

\subsection{Data-driven learning methods} 

When sufficient data is available, many works have investigated neural networks for learning the dynamics governed by time-dependent PDEs. These methods generally fall into two categories. The first category leverages well-established architectures from the computer vision domain, such as classical convolutional neural networks (CNNs) \cite{stachenfeld2021learned,bar2019learning}, U-Net \cite{guptatowards}, ResNet \cite{lu2018beyond}, and ConvLSTM \cite{shi2015convolutional}, to capture intricate spatiotemporal patterns. These architectures excel at extracting local features and are naturally suited for data represented on regular grids or meshes.
The second category involves neural operators, which focus on learning mappings between function spaces. This class includes models like DeepONet \cite{lu2021learning} and its variations \cite{seidman2022nomad,venturi2023svd,lee2023hyperdeeponet}, as well as the FNO \cite{li2020fourier} and its derivatives \cite{tran2023factorized,rahman2022u,wen2022u,zhang2024deciphering}. Neural operators offer the advantage of mesh independence, enabling generalization across different discretizations of the same class of PDEs. Additionally, advances in other data-driven neural PDE solvers, such as graph neural networks (GNNs) \cite{sanchez2020learning,pfaff2020learning} and transformer-based architectures \cite{wu2024transolver,hang2024unisolver,janny2023eagle,li2024scalable}, provide powerful tools to extend neural operators to various complex geometries.

Despite the progress achieved by these data-driven methods, they still face several limitations. One significant drawback is their heavy reliance on large and high-quality datasets, which are often time-consuming to generate or infeasible to many scientific and engineering domains. Moreover, these models are prone to overfitting and fail to generalize to unseen data, which is a key challenge in complex physical systems due to their inherent complexity and diversity.

\subsection{Physics-aware learning methods}

When data is limited, integrating physical knowledge into neural networks becomes important. This integration typically follows two strategies: physics-informed (soft constraints) and physics-encoded (hard constraints). Physics-informed methods incorporate physical laws, such as the PDE formulations, into the loss function. Representative works include PINN \cite{raissi2019physics}, PINO \cite{li2024physics}, PhyCRNet \cite{ren2022phycrnet}, MCNP \cite{zhang2025monte}, and PI-DeepONet\cite{wang2021learning}. These methods can be data-efficient but face challenges such as hyperparameter tuning, training instability, and requiring substantial prior knowledge of the governing equations \cite{krishnapriyan2021characterizing}.

On the other hand, physics-encoded approaches directly embed physical constraints into the network architecture, such as PDE operators, conservation laws, and boundary conditions \cite{long2019pde,Rao_2023,mi2024spatiotemporal,zeng2024phympgn}. These methods can be highly effective when working with limited data and incomplete physics priors. However, hard-constrained methods like PeRCNN, which use specially designed convolutional kernels to approximate local differential operators, often struggle to capture global information, limiting their performance on problems that require long-range spatial dependencies.

\begin{figure*}[!t] % 尝试 ht 或 h
    \centering
    \includegraphics[width=0.98\textwidth, keepaspectratio]{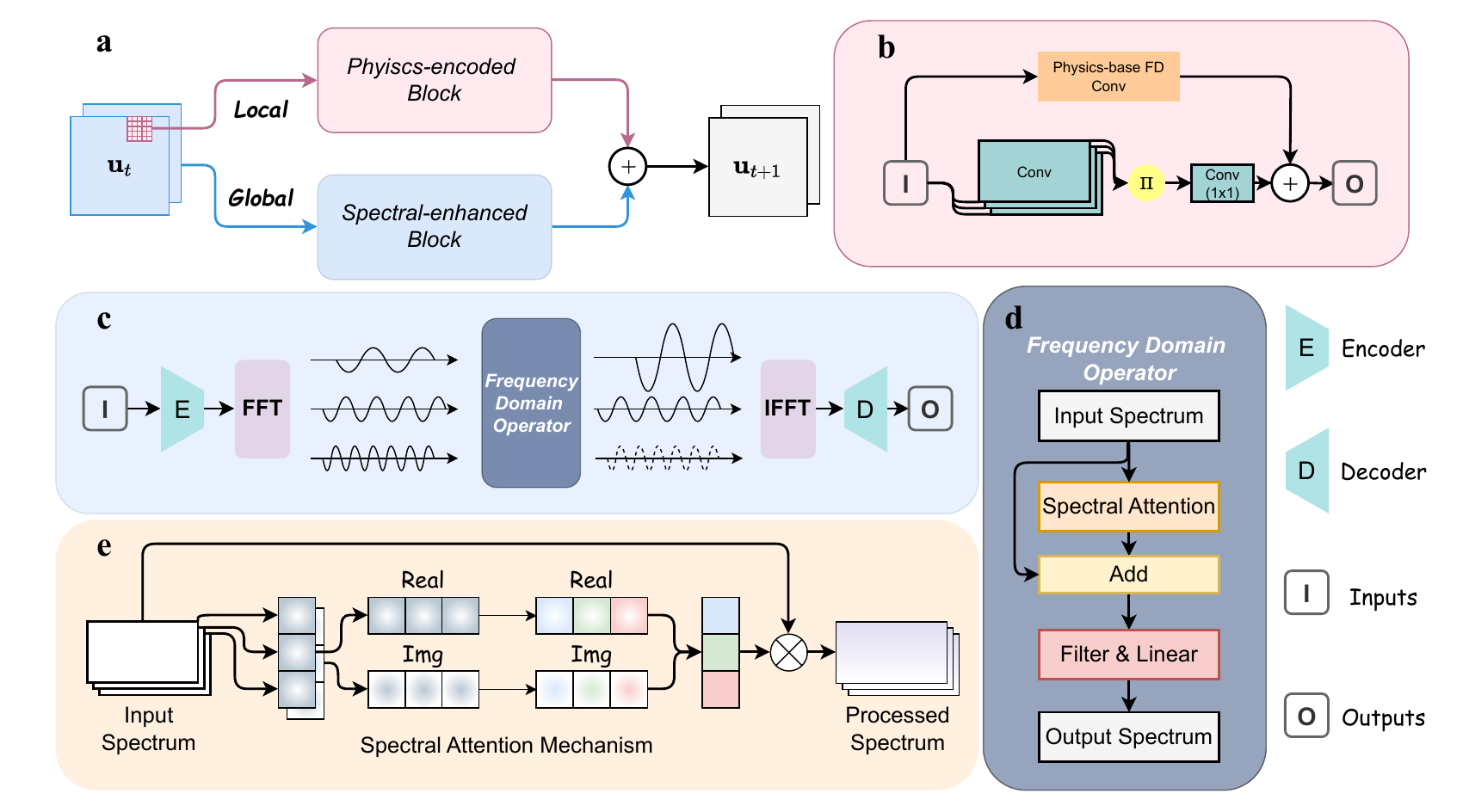} % 调整宽度
    \caption{\textbf{The architecture of PeSANet.} (a) Core blocks: physics-encoded and spectral-enhanced blocks to learn local and global features, respectively. (b) Physics-based convolution for known PDE terms (such as $\nabla ^2 \mathbf{u}$) and $\Pi$-block for unknown terms. (c) Spectral-enhanced block: including the FFT, encoder, frequency domain operator, IFFT, and decoder. (d) Frequency domain operator. (e) Spectral attention mechanism.}
    \label{fig:Main_Model}
\end{figure*}

\subsection{Hybrid learning methods}

Hybrid models provide an effective acceleration strategy for solving PDEs when the governing equations are fully known. These models couple neural networks with traditional numerical methods (e.g., FDM, FVM) to combine their advantages. Neural networks approximate computationally expensive components of the PDE solution, such as complex nonlinearities or fine-scale features, while established numerical methods handle the more tractable parts. This coupling enhances computational efficiency without sacrificing accuracy. Examples include multi-scale hybrid models \cite{vlachas2022multiscale} and other couplings with FVM (learned interpolation model \cite{kochkov2021machine} and temporal stencil modeling \cite{sun2023neural}) and FDM (P$^2$C$^2$Net \cite{wang2024p}). However, these models require complete knowledge of the PDEs, restricting their application to systems with unknown or partially unknown physics.

\section{Methodology}

\subsection{Problem Formulation }
Let us consider spatiotemporal complex systems governed by the following time-dependent PDEs:
 \begin{equation} \frac{\partial \mathbf{u}}{\partial t} = \mathcal{F} (t, \mathbf{x},  \mathbf{u}, \nabla \mathbf{u},\mathbf{u} \cdot \nabla \mathbf{u} ,\nabla ^2 \mathbf{u},\cdot\cdot\cdot ; \boldsymbol{\mu}), \label{pde} \end{equation}
where $\mathcal{F}(\cdot)$ represents an unknown operator that involves spatial derivatives of various orders, such as the gradient $\nabla \mathbf{u}$ and the Laplacian $\nabla ^2 \mathbf{u}$. The parameter $\boldsymbol{\mu}$ governs the physical properties of the system. Additionally, the initial conditions (ICs) and boundary conditions (BCs) are denoted as $\mathcal{I}[\mathbf{u}](\mathbf{x},t = 0) = 0$ and $\mathcal{B}[\mathbf{u}](\mathbf{x},t) = 0$, respectively.

In this paper, we aim to use the neural network as a surrogate model to simulate the complex systems based on incomplete physical prior of $\mathcal{F}$ and limited training data (e.g., 2-5 trajectories). The trained model is expected to learn the underlying dynamics of the physical system, which can accurately predict long-term system dynamics and generalize across different ICs. Traditional numerical and hybrid methods are not applicable to this scenario because they require sufficient prior knowledge of the physical system. On the other hand, data-driven methods demand large numbers of training data to achieve reasonable generalization performance. To this end, we propose the PeSANet.

\subsection{PeSANet Architecture}

In this section, we introduce the overall architecture of the proposed PeSANet (Figure \ref{fig:Main_Model}(a)). The model has two key components: the physics-encoded block and the spectral-enhanced block, which are designed to learn the local and global features of the complex system, respectively. The physics-encoded block uses multiplicative convolution to approximate the finite difference (FD) stencil, which extracts the local spatial information in a hard-constrained manner. The spectral-enhanced block conducts spectral attention operations in the frequency domain, which can further enhance the performance via capturing global features, such as long-range spatial dependencies. This synergistic combination allows the model to learn local details and global structures of the complex system, resulting in a thorough understanding of the underlying physical phenomena.

\subsection{Physics-encoded Block}

The physical-encoded block (Figure \ref{fig:Main_Model}(b)) aims to learn the underlying local differential operators that govern the complex system via a hard-constrained approach. Similar to PeRCNN \cite{Rao_2023}, we adopt the $\Pi$-block and physics-based convolutional (PyConv) layer to handle known and unknown operators, respectively. 

Considering the forward Euler scheme, the updated rule of the system in Equation \ref{pde} is given by:
\begin{equation}
    \mathbf{u}_{(k+1)\delta t} = {\mathcal{F}}(\mathbf{u}_{k\delta t}) \cdot \delta t +\mathbf{u}_{k\delta t},
\end{equation}
where $\delta t$ denotes the step size. The operator $\mathcal{F}$ is then approximated using the following $\Pi$-block $\hat{\mathcal{F}}$:
\begin{equation}
\label{percnn}
    \hat{\mathcal{F}}(\mathbf{u}_{k\delta t}) = \sum^{N_c}_{c=1} W_c \cdot  \left[ \prod^{N_l}_{l=1}(K_{c,l} \star \hat{\mathbf{u}}_{k\delta t}+b_l)\right],
\end{equation}
where $N_c$ represents the number of channels, and $N_l$ is the total number of parallel convolutional layers. The symbol $\star$ denotes the convolution operation. For each layer $l$ and channel $c$, $K_{c,l}$ denotes the filter weight, while $b_l$ represents the bias for layer $l$. $W_c$ is a $1 \times 1$ convolutional layer assigned to the $c^{\text{th}}$ channel. 

The main idea of the $\Pi$-block is to use multiplicative convolutional layers to approximate the FD stencil. Because of its multiplicative structure, the $\Pi$-block can efficiently approximate unknown nonlinear terms, such as $\mathbf{u} \cdot \nabla \mathbf{u}$. Theorem 1 in \cite{Rao_2023} demonstrates that the $\Pi$-block can serve as a universal polynomial approximator for the unknown nonlinear operator.

Furthermore, to effectively incorporate the partially known knowledge of the PDE system, we use the physics-based convolutional (PyConv) layer as a complementary strategy. When a term in the governing PDE is known, such as the Laplacian term \(\nabla^2 \mathbf{u}\), we directly set the kernel of the PyConv to align with the corresponding FD stencil. As a result, the PyConv layer is constructed to incorporate known physical information, while the $\Pi$-block is designed to learn the unknown operators. The combination of the $\Pi$-block and PyConv forms the physics-encoded block, which fully utilizes the physics structure of the differential operator and can efficiently model nonlinear systems from limited training data.

\subsection{Spectral-enhanced Block}

Although the physics-encoded block can efficiently approximate local operators within the complex system, it still fails to capture long-range spatial dependencies due to its limited receptive field. To overcome this limitation, we propose the spectral-enhanced block to process global information in the frequency domain, enabling a more comprehensive and holistic understanding of the complex system.

The proposed spectral-enhanced block (Figure \ref{fig:Main_Model}(c)) contains the following components. An encoder (E) is first employed to map the spatial physical field into the frequency domain via the Fast Fourier Transform (FFT). After that, a frequency domain operator processes the transformed information in the frequency domain. Subsequently, the data is returned to the spatial domain using an inverse FFT (IFFT), followed by a decoder (D) to reconstruct the spatial field.
% \paragraph{Fourier Modes Block}...

\subsubsection{Frequency Domain Operator}
The frequency domain operator (Figure \ref{fig:Main_Model}(d)) is specifically designed to process information in the frequency domain. The block takes in the input spectrum and applies the following series of operations.

Firstly, a spectral attention mechanism is introduced to capture inter-spectrum relationships, as detailed in the next section. This mechanism allows the model to identify and emphasize key spectral features critical for the task, enhancing its ability to capture dependencies across different spectral layers. Secondly, the processed spectrum is aggregated with the original spectrum via element-wise addition. This skip connection strategy helps ensure robustness during training and preserves the original frequency information. Finally, the aggregated spectrum undergoes filtering and a linear transformation, aligning with the traditional neural Fourier layer \cite{li2020fourier}.

These operations work together to handle the frequency signals, enhance the model's representational capacity, and ultimately produce the output spectrum.

\subsubsection{Spectral Attention Mechanism}

Motivated by the channel attention module \cite{wang2020eca}, we introduce the spectral attention mechanism (Figure \ref{fig:Main_Model}(e)) within the frequency domain operator, which effectively exploits the inter-spectrum relationships among spectral features. Given a complex feature denoted as $\mathbf{X} + i\mathbf{Y} \in \mathbb{C}^{c\times k_1 \times k_2}$, where $c$ denotes the number of channels and $k_1 \times k_2$ represents the number of frequency components, we first aggregate the frequency information of each feature map via the following average-pooling ($\operatorname{Avg}$) and max-pooling ($\operatorname{Max}$) operations:
\begin{equation}
    \begin{aligned}
(\mathbf{X}_{\operatorname{avg}}, \mathbf{Y}_{\operatorname{avg}}) &=  (\operatorname{Avg}(\mathbf{X}), \operatorname{Avg}(\mathbf{Y}) ) \in \mathbb{R}^{c} \times \mathbb{R}^{c}, \\
(\mathbf{X}_{\operatorname{max}}, \mathbf{Y}_{\operatorname{max}}) &= ( \operatorname{Max}(\mathbf{X}), \operatorname{Max}(\mathbf{Y}) ) \in \mathbb{R}^{c}\times \mathbb{R}^{c}.
    \end{aligned}
\end{equation}

We handle the real and imaginary parts separately because the complex field is not well-ordered, i.e., there is no natural ordering between complex numbers. As a result, directly applying the max-pooling operation to the complex-valued features would lead to meaningless results. After the pooling operations, we apply four multi-layer perceptrons (MLPs) with one hidden layer to capture the inter-spectrum relationships as follows:
\begin{equation}
    \begin{aligned}
{\operatorname{att}}(\mathbf{X}) &= \sigma\left( \operatorname{MLP}_{\operatorname{1}}(\mathbf{X}_{\operatorname{avg}}) + \operatorname{MLP}_{\operatorname{2}}(\mathbf{X}_{\operatorname{max}}) \right) \in \mathbb{R}^{c}, \\
{\operatorname{att}}(\mathbf{Y}) &= \sigma\left( \operatorname{MLP}_{\operatorname{3}}(\mathbf{Y}_{\operatorname{avg}}) + \operatorname{MLP}_{\operatorname{4}}(\mathbf{Y}_{\operatorname{max}}) \right) \in \mathbb{R}^{c}, \label{eq_att}
    \end{aligned}
\end{equation}
where $\sigma$ represents an activation function. Applying the attention coefficients to re-weight the input spectrum, the processed spectrum $\mathbf{X}_{\operatorname{p}} + i\mathbf{Y}_{\operatorname{p}} \in \mathbb{C}^{c\times k_1 \times k_2}$ can be obtained as:
\begin{equation}
        \mathbf{X}_{\operatorname{p}} + i\mathbf{Y}_{\operatorname{p}} = (\mathbf{X} + i\mathbf{Y}) \otimes (\operatorname{att}(\mathbf{X}) + i \operatorname{att}(\mathbf{Y})),  \label{x_p}
\end{equation}
where $\otimes$ denotes the element-wise multiplication.

\begin{table*}[!ht]
\centering

\begin{tabular}{ c c c c c c }
\toprule
Case & Numerical Methods & Spatial Domain & Temporal Domain & Training Trajectories & Test Trajectories \\
\midrule

Burgers & FDM & $(0,1)^2$ & $[0,3.216]$ & 5 & 5 \\
% \hline
FN & FDM & $(0,128)^2$ & $[0,10]$ & 5 & 5 \\
% \hline
GS & FDM & $(0,1)^2$ & $[0,1900]$ & 2 & 5 \\
% \hline
NSE & FVM &$(0, 2 \pi)^2$ & $[0,10]$ & 5 & 14 \\
\bottomrule
\end{tabular}
\caption{\textbf{Summary of experimental settings for different cases.}}
\label{tab:data}
 
\end{table*}

Compared to the traditional neural Fourier layer \cite{li2020fourier}, the spectral attention mechanism further enhances the representation capability for the following reasons. On the one hand, the spectral attention mechanism uses all frequency components to calculate the attention coefficients, instead of directly truncating high-frequency signals. This comprehensive consideration ensures all critical frequency information in the spectral domain is leveraged effectively. On the other hand, the attention mechanism in Equation \ref{eq_att} enhances the interactions between spectral layers by dynamically capturing their internal relationship. This interaction allows information to be processed jointly across spectrum layers, resulting in a more expressive representation of spectral features.

\begin{table}[!t]
% \vspace{-20pt}
\centering

\resizebox{0.482\textwidth}{!}
{\begin{tabular}{ccccc}
\toprule
Case & Model & RMSE $\downarrow$& MAE $\downarrow$& HCT (s) $\uparrow$\\ \midrule
 
\multirow{5}{*}{Burgers} 
% & DeepONet & 5.8507 & 7.6036 & 9.6\\ 
& FactFormer & 0.0142 & 0.0109 & 3.216 \\ 
& FNO & 0.1565 & 0.1301 & 0.104 \\ 
& F-FNO & 0.1097 & 0.0877 & 0.088 \\ 
& PeRCNN & \underline{0.0075}& \underline{0.0058} & \underline{3.216} \\ 
\cmidrule{2-5}
& PeSANet (Ours) & $\mathbf{0.0069}$  & $\mathbf{0.0051}$  & $\mathbf{3.216}$ \\ 
& Promotion ($\uparrow$) & \textcolor{blue}{8.00\%} & \textcolor{blue}{12.07\%} & \textcolor{blue}{0.00\%$^{*}$}   \\      
\midrule
\multirow{5}{*}{FN} 
% & DeepONet & 5.8507 & 7.6036 & 9.6\\ 
& FactFormer & 0.8664 & 0.3472 & 1.62 \\ 
& FNO & 3196737 & 1892978 & 1.77 \\ 
& F-FNO & 0.5328 & 0.3828 & 0.06 \\ 
& PeRCNN & \underline{0.2281}& \underline{0.1526} & \underline{6.27} \\ 
\cmidrule{2-5}
& PeSANet (Ours) & $\mathbf{0.1577}$  & $\mathbf{0.0784}$  & $\mathbf{8.13}$ \\ 
& Promotion ($\uparrow$) & \textcolor{blue}{30.86\%} & \textcolor{blue}{48.62\%} & \textcolor{blue}{29.66\%}   \\     
\midrule
\multirow{5}{*}{GS} 
% & DeepONet & 5.8507 & 7.6036 & 9.6\\ 
& FactFormer & 0.1095 & 0.0704 & 982.5 \\ 
& FNO & NaN & NaN & 330.0 \\ 
& F-FNO & 0.1265 & 0.0829 & 832.5 \\ 
& PeRCNN & \underline{0.0993}& \underline{0.0595} & \underline{1110.0} \\ 
\cmidrule{2-5}
& PeSANet (Ours) & $\mathbf{0.0481}$  & $\mathbf{0.0307}$  & $\mathbf{1327.5}$ \\ 
& Promotion ($\uparrow$) & \textcolor{blue}{51.56\%} & \textcolor{blue}{48.4\%} & \textcolor{blue}{19.59\%}   \\     
\midrule
\multirow{5}{*}{NSE} 
& FactFormer & 0.8971 & 0.6032 & 0.896 \\ 
& FNO & 0.8625 & 0.6617 & 2.464 \\ 
& F-FNO & 0.6992 & 0.5171 & 3.808 \\ 
& PeRCNN & \underline{0.3533}& \underline{0.1993} & \underline{7.197} \\ 
\cmidrule{2-5}
& PeSANet (Ours) & $\mathbf{0.2308}$  & $\mathbf{0.1573}$  & $\mathbf{7.728}$ \\ 
& Promotion ($\uparrow$) & \textcolor{blue}{34.67\%} & \textcolor{blue}{21.07\%} & \textcolor{blue}{7.37\%}   \\      
\bottomrule
% \footnote{}
% 这里能这么写吗？$*$ meaning that the predicting last to the end
\end{tabular}}
% \vspace{10pt}
\caption{\textbf{Quantitative results of PeSANet and baselines.} The best results are in bold, and the second-best results are underlined. The $*$ indicates the inference process has reached the end of the test data.}
\label{tab:model_performance}
\end{table}
\section{Experiments}

\begin{figure*}[!ht] % 尝试 ht 或 h
    \centering
    \includegraphics[width=1\textwidth, keepaspectratio]{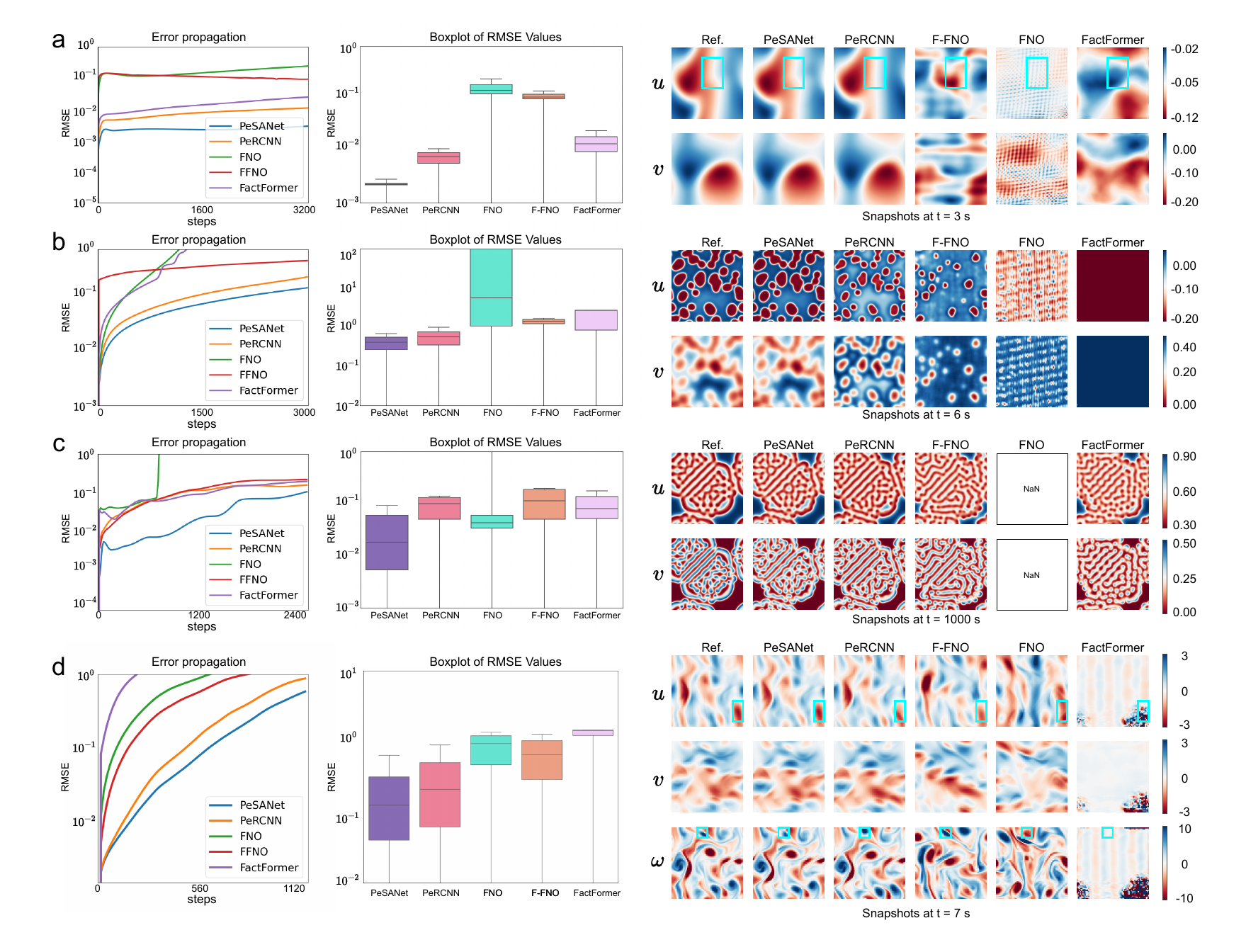} % 调整宽度
    \vspace{-20pt}
    \caption{\textbf{An overview comparison between PeSANet and other baselines:} error propagation curves (left), error boxplots (middle), and final prediction plots (right). Figures (a-d) respectively show the qualitative results for the 2D Burgers, 2D FN, 2D GS, and 2D NSE cases.}
    \label{fig:Main_Result}
\end{figure*}

To rigorously evaluate the efficacy and generalizability of the proposed PeSANet, we conducted a series of comprehensive experiments on a wide range of complex systems, such as fluid dynamics and reaction-diffusion systems. Specifically, the model is evaluated on the two-dimensional Burgers' equation, the two-dimensional FitzHugh-Nagumo (FN) system, the two-dimensional Gray-Scott (GS) system, and the two-dimensional Navier-Stokes equations (NSE). The experimental results demonstrate that the PeSANet consistently achieves superior accuracy compared to existing methodologies across all challenging benchmarks. These experiments not only showcase the model's robust adaptability to a diverse set of systems but also confirm its ability to accurately predict long-term dynamics.

\subsection{Experimental Setup}
This section details the experimental setup, including the datasets, baselines, metrics, and the training procedure.

\textbf{Datasets.} We considered several two-dimensional PDE-governed nonlinear complex systems, including the Burgers' equation, the FitzHugh-Nagumo (FN) system, the Gray-Scott (GS) system, and the Navier-Stokes equations (NSE). These systems represent diverse physical phenomena and dynamic behaviors, providing a comprehensive assessment of the model's performance and generalization capabilities. In this paper, we target the data-scare scenario, and each experiment includes 2-5 trajectories in the training set. We summarize the datasets in Table \ref{tab:data}, with a more detailed introduction in Appendix Dataset Informations.

\textbf{Baseline Models.} To demonstrate the effectiveness of PeSANet, we compared it with the following baseline models:
\begin{itemize} \item \textbf{FNO} \cite{li2020fourier}: A widely recognized high-performing data-driven neural operator, which utilizes the Fourier transform to handle PDE information in the frequency domain. 
\item \textbf{PeRCNN} \cite{Rao_2023}: A model that embeds physical knowledge in a hard-constrained manner and demonstrates excellent long-term prediction ability.
\item \textbf{F-FNO} \cite{tran2023factorized}: The Factorized Fourier Neural Operator, which employs a separable Fourier layer and enhanced residual connections to effectively reduce computational complexity and improve scalability.
\item \textbf{FactFormer} \cite{li2024scalable}: A model that addresses the computational expense and numerical instability of applying Transformers to PDEs with large grids by employing an axial factorized kernel integral. \end{itemize}

\textbf{Evaluation Metrics.} We use three types of metrics to quantitatively assess the model performance: Root Mean Square Error (RMSE), Mean Absolute Error (MAE), and High Correction Time (HCT). Specifically, RMSE quantifies the average magnitude of prediction errors in the original data units, offering a general measure of model performance. MAE, conversely, measures the average absolute difference between predicted and ground truth values, exhibiting robustness to outliers compared to RMSE. Finally, HCT quantifies the temporal duration required for the model's predictions to converge to a high level of accuracy, a critical consideration for real-time applications. The complete mathematical formulations of these metrics can be found in Appendix Evaluation Metrics.

\textbf{Model Training.} We train our model autoregressively, where the model predicts an output at each iteration, and the output is used as input for the next prediction. The mean squared error loss function is utilized for training, optimized using the Adam optimizer.

\subsection{Main Results}

The performance characteristics of the proposed PeSANet are comprehensively evaluated and presented in Figure \ref{fig:Main_Result} and Table \ref{tab:model_performance}. Demonstrating robust performance despite limited training data, Figure \ref{fig:Main_Result} provides visualizations of error distribution, error propagation characteristics, and predicted trajectories. These qualitative results are further substantiated by the quantitative metrics presented in Table \ref{tab:model_performance}. 

\textbf{2D Burgers Equation.} The efficacy of the proposed PeSANet framework is demonstrated in the context of the 2D Burgers equation (Figure \ref{fig:Main_Result}(a)). While PeRCNN exhibited some success in the beginning, it struggled with long-term predictions, as evidenced by the errors observed in Figure \ref{fig:Main_Result}(a). In contrast, the PeSANet framework demonstrates superior performance in capturing the underlying physical dynamics, particularly in long-term forecasting. This qualitative advantage is further corroborated by the quantitative metrics presented in Table \ref{tab:model_performance}, which show significant improvements in both RMSE and MAE. The divergence of PeRCNN is also evident in the final time step snapshot.

\textbf{2D FitzHugh-Nagumo Equation.} Figure \ref{fig:Main_Result}(b) demonstrates the difficulties encountered by data-driven methods (F-FNO, FNO, and FactFormer) when trained on limited data, with FactFormer showing near-complete failure. Even F-FNO, the strongest among these, exhibits limitations in the long-term prediction of the FN dynamics. Although PeRCNN, incorporating physics-informed structures, generates predictions with similar contours to the ground truth, its long-term accuracy remains limited. In contrast, our model achieves at least a 30\% improvement over the best baseline, demonstrating the benefits of our approach for small datasets and the crucial role of physics-encoded block in enhancing long-term prediction accuracy.

\textbf{2D Gray-Scott Equation.} According to  Table \ref{tab:model_performance}, PeSANet achieves substantial gains of approximately 50\% in both RMSE and MAE. The complex, labyrinthine patterns of the GS-RD system are visualized in the snapshots presented in Figure \ref{fig:Main_Result} (c). A clear visual comparison reveals that PeSANet maintains a reasonable resemblance to the ground truth even after long-term prediction, with only minor deviations. In contrast, other baseline models fail to capture the intricate patterns or even the basic contours of the structures.

\textbf{2D Navier-Stokes Equation.} For complex fluid dynamics, we evaluate the performance of the proposed PeSANet with other baseline models on a Navier-Stokes Equations (NSE) dataset with a Reynolds number of 1000. This task requires models that can accurately capture both local variations and global features to ensure reliable long-term predictions. Compared to other baseline, our model achieves significant improvements of 20\% and 30\% in RMSE and MAE (Table \ref{tab:model_performance}). The error propagation curves in Figure \ref{fig:Main_Result}(d) further validate the superiority of PeSANet, with its error consistently lower than that of other models. The snapshot at $t = 7 s$ shows near-perfect agreement between PeSANet’s predictions and the ground truth, while other models exhibit significant deviations. This collective evidence establishes PeSANet as the superior and more robust model for long-term prediction tasks in complex fluid dynamics systems.

\subsection{Generalization Test}
\begin{figure}[!t]
     \centering
     \includegraphics[width=0.98\linewidth]{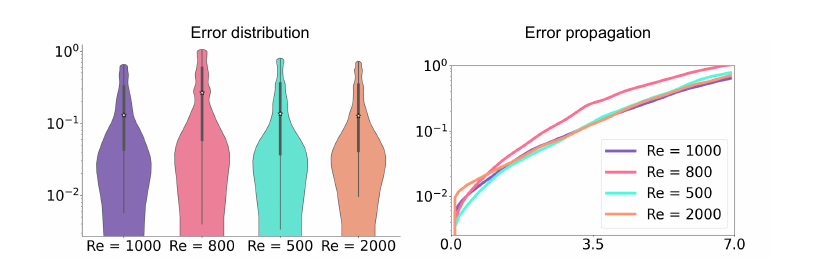}
     \caption{\textbf{Generalization test.} The error distribution and propagation of PeSANet for generalization over different Reynolds numbers.}
     \label{fig:enter-label}
     % \vspace{5pt}
 \end{figure}

Our model leverages incorporated physical knowledge to enable effective fine-tuning and robust generalization with scarce training data. Using the NSE as a representative example, PeSANet demonstrates excellent performance at the training Reynolds number (Re = 1000) and further exhibits strong generalization capabilities by fine-tuning with limited data to new Reynolds numbers (Re = 500, 800, and 2000). The error distributions across these Reynolds numbers (Figure \ref{fig:enter-label}) confirm the generalization ability of PeSANet. The error curves maintain similar characteristics to the Re = 1000 case, with error propagation well controlled, demonstrating the model's reliability across different flow regimes. It is worth mentioning that fine-tuning with only two data samples enables PeSANet to generate predictions closely approximating the ground truth, further validating its robustness and generalization capacity. In summary, PeSANet effectively captures the underlying physical principles of the system, achieving accurate predictions for NSE at varying Reynolds numbers with minimal fine-tuning data.

\subsection{Ablation Study}

To evaluate the contributions of different components in our PeSANet, we designed a series of ablation studies based on the FN equation. Specifically, we conducted experiments by (1) PeSANet w/o SA, removing the spectral attention component, (2) PeSANet w/o Pe, removing the physical-encoded block in the physical module, (3) Pe+FNO, replacing the spectral-enhanced block with FNO, and (4) Pe+F-FNO, replacing the spectral-enhanced block with F-FNO.

\begin{table}[!t]
  \centering
  % \vspace{3pt}
  \begin{tabular}{cccc}
    \toprule
    Model & RMSE $\downarrow$& MAE $\downarrow$& HCT (s) $\uparrow$ \\
    \midrule
    PeSANet          & 0.1577 & 0.0784 & 8.13 \\
    PeSANet w/o SA   & 0.1980 & 0.1073 & 7.65\\
    PeSANet w/o Pe   & NaN    & NaN & 0.18   \\
    Pe+FNO      & NaN    & NaN & 0.002   \\
    Pe+F-FNO      & 0.5118    & 0.3432  & 2.25 \\
    \bottomrule
  \end{tabular}
\caption{\textbf{Results for ablation study on the FN equation.}}
  \label{tab:model-ab}
  % \vspace{5pt}
\end{table}

As shown in Table \ref{tab:model-ab}, experimental results demonstrate that removing the spectral attention module leads to a significant decline in network performance, highlighting the crucial role of the spectral attention mechanism in capturing spectral information and enhancing the model's representational capacity. The removal of the physics-encoded block results in poor performance on small datasets, indicating the importance of prior physical knowledge in improving the model's generalization ability, especially when data is limited. Furthermore, completely replacing the spectral-enhanced block with FNO or F-FNO also results in a significant performance drop. 

In conclusion, these experimental results demonstrate that physical embedding and spectral enhancement are indispensable components of the PeSANet framework, providing the model with powerful representational ability and robustness.

\section{Conclusion}

This paper introduces PeSANet, a predictive model designed to enhance the prediction accuracy of complex systems governed by PDEs by integrating global and local information, leveraging available physical data, and emphasizing frequency-domain feature processing. The PeSANet is designed to not only capture the overall system dynamics but also exploit local information through hard-coding physical knowledge. Furthermore, it incorporates a carefully designed spectral attention module to capture key spectral features, effectively handling the global information in the frequency domain. These strategies significantly improve PeSANet's data efficiency, enabling superior long-term prediction performance and generalization capabilities even with limited training data. Experimental results demonstrate that PeSANet achieves state-of-the-art predictive performance on four distinct nonlinear dynamical systems with small datasets, exhibiting strong adaptability and flexibility even with limited data and varying initial conditions. The generalization test also demonstrates PeSANet's transfer learning ability. In conclusion, PeSANet offers a novel approach to long-term forecasting by integrating physics and spectral attention, demonstrating superior performance in data-scarce scenarios.

\newpage
\section*{Acknowledgments}
The work is supported by the National Natural Science Foundation of China (No. 62276269, No. 92270118), the Beijing Natural Science Foundation (No. 1232009), and the Strategic Priority Research Program of the Chinese Academy of Sciences (No. XDB0620103). Hao Sun and Yang Liu would like to thank the support from the Fundamental Research Funds for the Central Universities (No. 202230265, No. E2EG2202X2).
\bibliographystyle{named}
\bibliography{ijcai25}

\begin{thebibliography}{}

\bibitem[\protect\citeauthoryear{Anderson and Wendt}{1995}]{anderson1995computational}
John~David Anderson and John Wendt.
\newblock {\em Computational Fluid Dynamics}, volume 206.
\newblock Springer, 1995.

\bibitem[\protect\citeauthoryear{Bar-Sinai \bgroup \em et al.\egroup }{2019}]{bar2019learning}
Yohai Bar-Sinai, Stephan Hoyer, Jason Hickey, and Michael~P Brenner.
\newblock Learning data-driven discretizations for partial differential equations.
\newblock {\em Proceedings of the National Academy of Sciences}, 116(31):15344--15349, 2019.

\bibitem[\protect\citeauthoryear{Ghergu and Radulescu}{2011}]{ghergu2011nonlinear}
Marius Ghergu and Vicentiu Radulescu.
\newblock {\em Nonlinear PDEs: Mathematical models in biology, chemistry and population genetics}.
\newblock Springer Science \& Business Media, 2011.

\bibitem[\protect\citeauthoryear{Gupta and Brandstetter}{2023}]{guptatowards}
Jayesh~K Gupta and Johannes Brandstetter.
\newblock Towards multi-spatiotemporal-scale generalized pde modeling.
\newblock {\em Transactions on Machine Learning Research}, 2023.

\bibitem[\protect\citeauthoryear{Hang \bgroup \em et al.\egroup }{2024}]{hang2024unisolver}
Zhou Hang, Yuezhou Ma, Haixu Wu, Haowen Wang, and Mingsheng Long.
\newblock Unisolver: Pde-conditional transformers are universal pde solvers.
\newblock {\em arXiv preprint arXiv:2405.17527}, 2024.

\bibitem[\protect\citeauthoryear{Holton and Hakim}{2013}]{holton2013introduction}
James~R Holton and Gregory~J Hakim.
\newblock {\em An introduction to dynamic meteorology}, volume~88.
\newblock Academic press, 2013.

\bibitem[\protect\citeauthoryear{Janny \bgroup \em et al.\egroup }{2023}]{janny2023eagle}
Steeven Janny, Aur{\'e}lien Beneteau, Madiha Nadri, Julie Digne, Nicolas Thome, and Christian Wolf.
\newblock Eagle: Large-scale learning of turbulent fluid dynamics with mesh transformers.
\newblock {\em arXiv preprint arXiv:2302.10803}, 2023.

\bibitem[\protect\citeauthoryear{Karniadakis and Sherwin}{2005}]{karniadakis2005spectral}
George Karniadakis and Spencer~J. Sherwin.
\newblock {\em Spectral/HP Element Methods for Computational Fluid Dynamics}.
\newblock Oxford University Press, USA, 2005.

\bibitem[\protect\citeauthoryear{Kochkov \bgroup \em et al.\egroup }{2021}]{kochkov2021machine}
Dmitrii Kochkov, Jamie~A Smith, Ayya Alieva, Qing Wang, Michael~P Brenner, and Stephan Hoyer.
\newblock Machine learning--accelerated computational fluid dynamics.
\newblock {\em Proceedings of the National Academy of Sciences}, 118(21):e2101784118, 2021.

\bibitem[\protect\citeauthoryear{Krishnapriyan \bgroup \em et al.\egroup }{2021}]{krishnapriyan2021characterizing}
Aditi Krishnapriyan, Amir Gholami, Shandian Zhe, Robert Kirby, and Michael~W Mahoney.
\newblock Characterizing possible failure modes in physics-informed neural networks.
\newblock {\em Advances in neural information processing systems}, 34:26548--26560, 2021.

\bibitem[\protect\citeauthoryear{Lee \bgroup \em et al.\egroup }{2023}]{lee2023hyperdeeponet}
Jae~Yong Lee, SungWoong CHO, and Hyung~Ju Hwang.
\newblock Hyperdeep{ON}et: learning operator with complex target function space using the limited resources via hypernetwork.
\newblock In {\em International Conference on Learning Representations}, 2023.

\bibitem[\protect\citeauthoryear{Li \bgroup \em et al.\egroup }{2021}]{li2020fourier}
Zongyi Li, Nikola Kovachki, Kamyar Azizzadenesheli, Burigede Liu, Kaushik Bhattacharya, Andrew Stuart, and Anima Anandkumar.
\newblock Fourier neural operator for parametric partial differential equations.
\newblock {\em International Conference on Learning Representations}, 2021.

\bibitem[\protect\citeauthoryear{Li \bgroup \em et al.\egroup }{2024a}]{li2024scalable}
Zijie Li, Dule Shu, and Amir Barati~Farimani.
\newblock Scalable transformer for pde surrogate modeling.
\newblock {\em Advances in Neural Information Processing Systems}, 36, 2024.

\bibitem[\protect\citeauthoryear{Li \bgroup \em et al.\egroup }{2024b}]{li2024physics}
Zongyi Li, Hongkai Zheng, Nikola Kovachki, David Jin, Haoxuan Chen, Burigede Liu, Kamyar Azizzadenesheli, and Anima Anandkumar.
\newblock Physics-informed neural operator for learning partial differential equations.
\newblock {\em ACM/JMS Journal of Data Science}, 1(3):1--27, 2024.

\bibitem[\protect\citeauthoryear{Long \bgroup \em et al.\egroup }{2019}]{long2019pde}
Zichao Long, Yiping Lu, and Bin Dong.
\newblock Pde-net 2.0: Learning pdes from data with a numeric-symbolic hybrid deep network.
\newblock {\em Journal of Computational Physics}, 399:108925, 2019.

\bibitem[\protect\citeauthoryear{Lu \bgroup \em et al.\egroup }{2018}]{lu2018beyond}
Yiping Lu, Aoxiao Zhong, Quanzheng Li, and Bin Dong.
\newblock Beyond finite layer neural networks: Bridging deep architectures and numerical differential equations.
\newblock In {\em International Conference on Machine Learning}, pages 3276--3285. PMLR, 2018.

\bibitem[\protect\citeauthoryear{Lu \bgroup \em et al.\egroup }{2021}]{lu2021learning}
Lu~Lu, Pengzhan Jin, Guofei Pang, Zhongqiang Zhang, and George~Em Karniadakis.
\newblock Learning nonlinear operators via deeponet based on the universal approximation theorem of operators.
\newblock {\em Nature machine intelligence}, 3(3):218--229, 2021.

\bibitem[\protect\citeauthoryear{Mi and Sun}{2024}]{mi2024spatiotemporal}
Yuan Mi and Hao Sun.
\newblock Spatiotemporal learning on cell-embedded graphs.
\newblock {\em arXiv preprint arXiv:2409.18013}, 2024.

\bibitem[\protect\citeauthoryear{Moukalled \bgroup \em et al.\egroup }{2016}]{darwish2016finite}
Fadl Moukalled, Luca Mangani, and Marwan Darwish.
\newblock {\em The finite volume method in computational fluid dynamics}.
\newblock Springer, 2016.

\bibitem[\protect\citeauthoryear{Pfaff \bgroup \em et al.\egroup }{2020}]{pfaff2020learning}
Tobias Pfaff, Meire Fortunato, Alvaro Sanchez-Gonzalez, and Peter~W Battaglia.
\newblock Learning mesh-based simulation with graph networks.
\newblock {\em arXiv preprint arXiv:2010.03409}, 2020.

\bibitem[\protect\citeauthoryear{Rahman \bgroup \em et al.\egroup }{2022}]{rahman2022u}
Md~Ashiqur Rahman, Zachary~E Ross, and Kamyar Azizzadenesheli.
\newblock U-no: U-shaped neural operators.
\newblock {\em arXiv preprint arXiv:2204.11127}, 2022.

\bibitem[\protect\citeauthoryear{Raissi \bgroup \em et al.\egroup }{2019}]{raissi2019physics}
Maziar Raissi, Paris Perdikaris, and George~E Karniadakis.
\newblock Physics-informed neural networks: A deep learning framework for solving forward and inverse problems involving nonlinear partial differential equations.
\newblock {\em Journal of Computational physics}, 378:686--707, 2019.

\bibitem[\protect\citeauthoryear{Rao \bgroup \em et al.\egroup }{2022}]{rao2022discovering}
Chengping Rao, Pu~Ren, Yang Liu, and Hao Sun.
\newblock Discovering nonlinear pdes from scarce data with physics-encoded learning.
\newblock {\em arXiv preprint arXiv:2201.12354}, 2022.

\bibitem[\protect\citeauthoryear{Rao \bgroup \em et al.\egroup }{2023}]{Rao_2023}
Chengping Rao, Pu~Ren, Qi~Wang, Oral Buyukozturk, Hao Sun, and Yang Liu.
\newblock Encoding physics to learn reaction--diffusion processes.
\newblock {\em Nature Machine Intelligence}, 5(7):765--779, 2023.

\bibitem[\protect\citeauthoryear{Ren \bgroup \em et al.\egroup }{2022}]{ren2022phycrnet}
Pu~Ren, Chengping Rao, Yang Liu, Jian-Xun Wang, and Hao Sun.
\newblock Phycrnet: Physics-informed convolutional-recurrent network for solving spatiotemporal pdes.
\newblock {\em Computer Methods in Applied Mechanics and Engineering}, 389:114399, 2022.

\bibitem[\protect\citeauthoryear{Sanchez-Gonzalez \bgroup \em et al.\egroup }{2020}]{sanchez2020learning}
Alvaro Sanchez-Gonzalez, Jonathan Godwin, Tobias Pfaff, Rex Ying, Jure Leskovec, and Peter Battaglia.
\newblock Learning to simulate complex physics with graph networks.
\newblock In {\em International conference on machine learning}, pages 8459--8468. PMLR, 2020.

\bibitem[\protect\citeauthoryear{Seidman \bgroup \em et al.\egroup }{2022}]{seidman2022nomad}
Jacob~H Seidman, Georgios Kissas, Paris Perdikaris, and George~J. Pappas.
\newblock {NOMAD}: Nonlinear manifold decoders for operator learning.
\newblock In Alice~H. Oh, Alekh Agarwal, Danielle Belgrave, and Kyunghyun Cho, editors, {\em Advances in Neural Information Processing Systems}, 2022.

\bibitem[\protect\citeauthoryear{Shi \bgroup \em et al.\egroup }{2015}]{shi2015convolutional}
Xingjian Shi, Zhourong Chen, Hao Wang, Dit-Yan Yeung, Wai-Kin Wong, and Wang-chun Woo.
\newblock Convolutional lstm network: A machine learning approach for precipitation nowcasting.
\newblock {\em Advances in neural information processing systems}, 28, 2015.

\bibitem[\protect\citeauthoryear{Stachenfeld \bgroup \em et al.\egroup }{2021}]{stachenfeld2021learned}
Kimberly Stachenfeld, Drummond~B Fielding, Dmitrii Kochkov, Miles Cranmer, Tobias Pfaff, Jonathan Godwin, Can Cui, Shirley Ho, Peter Battaglia, and Alvaro Sanchez-Gonzalez.
\newblock Learned coarse models for efficient turbulence simulation.
\newblock {\em arXiv preprint arXiv:2112.15275}, 2021.

\bibitem[\protect\citeauthoryear{Sun \bgroup \em et al.\egroup }{2023}]{sun2023neural}
Zhiqing Sun, Yiming Yang, and Shinjae Yoo.
\newblock A neural pde solver with temporal stencil modeling.
\newblock In {\em International Conference on Machine Learning}, pages 33135--33155. PMLR, 2023.

\bibitem[\protect\citeauthoryear{Tran \bgroup \em et al.\egroup }{2023}]{tran2023factorized}
Alasdair Tran, Alexander Mathews, Lexing Xie, and Cheng~Soon Ong.
\newblock Factorized fourier neural operators.
\newblock In {\em The Eleventh International Conference on Learning Representations}, 2023.

\bibitem[\protect\citeauthoryear{Venturi and Casey}{2023}]{venturi2023svd}
Simone Venturi and Tiernan Casey.
\newblock Svd perspectives for augmenting deeponet flexibility and interpretability.
\newblock {\em Computer Methods in Applied Mechanics and Engineering}, 403:115718, 2023.

\bibitem[\protect\citeauthoryear{Vlachas \bgroup \em et al.\egroup }{2022}]{vlachas2022multiscale}
Pantelis~R Vlachas, Georgios Arampatzis, Caroline Uhler, and Petros Koumoutsakos.
\newblock Multiscale simulations of complex systems by learning their effective dynamics.
\newblock {\em Nature Machine Intelligence}, 4(4):359--366, 2022.

\bibitem[\protect\citeauthoryear{Wang \bgroup \em et al.\egroup }{2020}]{wang2020eca}
Qilong Wang, Banggu Wu, Pengfei Zhu, Peihua Li, Wangmeng Zuo, and Qinghua Hu.
\newblock Eca-net: Efficient channel attention for deep convolutional neural networks.
\newblock In {\em Proceedings of the IEEE/CVF conference on computer vision and pattern recognition}, pages 11534--11542, 2020.

\bibitem[\protect\citeauthoryear{Wang \bgroup \em et al.\egroup }{2021}]{wang2021learning}
Sifan Wang, Hanwen Wang, and Paris Perdikaris.
\newblock Learning the solution operator of parametric partial differential equations with physics-informed deeponets.
\newblock {\em Science advances}, 7(40):eabi8605, 2021.

\bibitem[\protect\citeauthoryear{Wang \bgroup \em et al.\egroup }{2024}]{wang2024p}
Qi~Wang, Pu~Ren, Hao Zhou, Xin-Yang Liu, Zhiwen Deng, Yi~Zhang, Ruizhi Chengze, Hongsheng Liu, Zidong Wang, Jian-Xun Wang, et~al.
\newblock P2c2 net: Pde-preserved coarse correction network for efficient prediction of spatiotemporal dynamics.
\newblock {\em arXiv preprint arXiv:2411.00040}, 2024.

\bibitem[\protect\citeauthoryear{Wen \bgroup \em et al.\egroup }{2022}]{wen2022u}
Gege Wen, Zongyi Li, Kamyar Azizzadenesheli, Anima Anandkumar, and Sally~M Benson.
\newblock U-fno—an enhanced fourier neural operator-based deep-learning model for multiphase flow.
\newblock {\em Advances in Water Resources}, 163:104180, 2022.

\bibitem[\protect\citeauthoryear{Wu \bgroup \em et al.\egroup }{2024}]{wu2024transolver}
Haixu Wu, Huakun Luo, Haowen Wang, Jianmin Wang, and Mingsheng Long.
\newblock Transolver: A fast transformer solver for pdes on general geometries.
\newblock {\em arXiv preprint arXiv:2402.02366}, 2024.

\bibitem[\protect\citeauthoryear{Zeng \bgroup \em et al.\egroup }{2024}]{zeng2024phympgn}
Bocheng Zeng, Qi~Wang, Mengtao Yan, Yang Liu, Ruizhi Chengze, Yi~Zhang, Hongsheng Liu, Zidong Wang, and Hao Sun.
\newblock Phympgn: Physics-encoded message passing graph network for spatiotemporal pde systems.
\newblock {\em arXiv preprint arXiv:2410.01337}, 2024.

\bibitem[\protect\citeauthoryear{Zhang \bgroup \em et al.\egroup }{2024}]{zhang2024deciphering}
Rui Zhang, Qi~Meng, and Zhi-Ming Ma.
\newblock Deciphering and integrating invariants for neural operator learning with various physical mechanisms.
\newblock {\em National Science Review}, 11(4):nwad336, 2024.

\bibitem[\protect\citeauthoryear{Zhang \bgroup \em et al.\egroup }{2025}]{zhang2025monte}
Rui Zhang, Qi~Meng, Rongchan Zhu, Yue Wang, Wenlei Shi, Shihua Zhang, Zhi-Ming Ma, and Tie-Yan Liu.
\newblock Monte carlo neural pde solver for learning pdes via probabilistic representation.
\newblock {\em IEEE Transactions on Pattern Analysis and Machine Intelligence}, 2025.

\bibitem[\protect\citeauthoryear{Zienkiewicz \bgroup \em et al.\egroup }{2005}]{zienkiewicz2005finite}
Olek~C Zienkiewicz, Robert~L Taylor, and Jian~Z Zhu.
\newblock {\em The Finite Element Method: Its Basis and Fundamentals}.
\newblock Elsevier, 2005.

\end{thebibliography}

\newpage

This appendix provides explanations for several critical aspects, including model details, data
generation, and object recognition and tracking.

\section*{Appendix: Training Details}

\label{sectionc1}
To ensure consistent experimental conditions, all experiments were conducted on a single 80GB NVIDIA A100 GPU and an Intel(R) Xeon(R) Platinum 8380 CPU (2.30GHz, 64 cores). For brevity, we present only the hyperparameters that were modified in this study; all remaining hyperparameters adhered to the settings specified in the original publication.

\begin{table}[h!]
\centering
\begin{tabular}{l c c c}
% \hline
\toprule
\textbf{Case} & \textbf{Batchsize} & \textbf{Num of epochs} & \textbf{Learning rate} \\  
\midrule

Burgers       & 8                  & 5000 & $1 \times 10^{-4}$                  \\  
GS          & 8                  & 5000        & $5\times 10^{-4}$          \\  
FN            & 32                 & 8000        & $5\times 10^{-4}$          \\  
NSE          & 32                 & 8000          & $1\times 10^{-4}$        \\  
\bottomrule
\end{tabular}
\caption{Training Details of PeSANet.}
\label{tab:PIMRL Train}
\end{table}

To optimize training across different PDEs, a StepLR scheduler was utilized to implement a learning rate decay schedule in each case. The learning rate was reduced by a multiplicative factor at regular intervals: 0.985 every 20 training steps for the Burgers' equation, 0.985 every 200 steps for the Gray-Scott (GS) system, 0.98 every 50 steps for the FitzHugh-Nagumo (FN) system, and 0.96 every 200 steps for the Navier-Stokes Equations (NSE). This strategy ensured a gradual reduction of the learning rate throughout the training process for all PDEs.

%% The file named.bst is a bibliography style file for BibTeX 0.99c
\section*{Appendix: Baseline Models}

To ensure a fair and comprehensive comparison, our proposed method was evaluated against several established state-of-the-art (SOTA) and classical baseline models. Details of these models are provided below.

\textbf{Fourier Neural Operator (FNO).} 
The Fourier Neural Operator (FNO) \cite{li2020fourier} leverages the power of Fourier transforms in conjunction with neural networks to learn mappings between function spaces. This approach comprises two key components. First, Fourier transforms are applied to the system state variables, enabling the model to learn representations in the frequency domain. Subsequently, an inverse Fourier transform projects these learned representations back into the physical space. Second, convolutional operations are employed to process the system state variables, capturing complementary information not effectively captured by the frequency-domain learning. The final output is then obtained by combining the outputs of these two components.

\textbf{PeRCNN.} PeRCNN \cite{Rao_2023} is a physics-informed learning methodology that directly embeds physical laws into the neural network architecture. It employs multiple parallel convolutional neural networks (CNNs) and leverages feature map multiplication to simulate polynomial equations. This design choice enhances the model's extrapolation and generalization capabilities.

\textbf{F-FNO.} The Factorized Fourier Neural Operator (F-FNO) \cite{tran2023factorized} introduces a separable Fourier representation and improved residual connections within the Fourier Neural Operator framework, enabling the training of deeper networks and achieving enhanced performance in simulating PDEs.

 \textbf{FactFormer.} FactFormer \cite{li2024scalable} leverages an axial factorized kernel integral, implemented through a learnable projection operator that decomposes the input function into one-dimensional sub-functions. This factorization enables efficient computation of an instance-based kernel, making FactFormer a suitable low-rank surrogate for the full attention mechanism in multi-dimensional settings.

\section*{Appendix: Dataset Informations}
Initial conditions are generated randomly for each PDE system. Specifically, the Burgers' equation is initialized using a Gaussian distribution. The FitzHugh-Nagumo (FN) equation is initialized with Gaussian noise for a transient warm-up phase, after which time series data are extracted. For the Gray-Scott (GS) system, reactants are initially placed at random positions, followed by diffusion.

\textbf{2D Burgers Equation.} The 2D Burgers' equation is commonly employed as a benchmark model for comparing and evaluating different computational algorithms, and describes the complex interaction between nonlinear convection and diffusion processes in the way like:
\begin{align}  
\frac{\partial u}{\partial t} &= -u u_x - v u_y + \nu   \nabla^2 u , \\  
\frac{\partial v}{\partial t} &= -u v_x - v v_y + \nu \nabla^2 v  .  
\end{align}  
The $u$ and $v$ are the fluid velocities and $\nu$ denotes the viscosity coefficient. In this case, we choose $\nu = 0.005$, where $\delta t = 0.001s$.

\textbf{2D FitzHugh-Nagumo Equation.} The FitzHugh-Nagumo system can be described by the equation:
 \begin{align}
      \frac{\partial u}{\partial t} &= \mu_u \nabla^2 u+u-u^3-v+\alpha,\\
     \frac{\partial v}{\partial t}&=  \mu_v \nabla^2 v+(u-v)\beta.
 \end{align}
The coefficients $\alpha = 0.01$ and $\beta = 0.25$, governing the reaction process, take distinct values, while the diffusion coefficients are $\mu_u = 1$ and $\mu_v = 100$. In terms of time,  $\delta t = 0.002s$.

\textbf{2D Gray-Scott Equation.} The Gray-Scott equations describe the temporal and spatial variations of chemical concentrations in reaction-diffusion systems, which can be described by the equation:
 \begin{align}
\frac{\partial u}{\partial t} &= D_u \nabla^2 u - uv^2 + F(1-u) ,\\
\frac{\partial v}{\partial t} &= D_v \nabla^2 v + uv^2 - (F+k)v .
 \end{align}
Here, in the two-dimensional case, $D_u$ and $D_v$ represent the diffusion coefficients of the two substances, with specific values of $D_u = 2.0 \times 10^{-5}$ and $D_v = 5.0 \times 10^{-6}$. $F = 0.04$ denotes the growth rate of the substance, while $k= 0.06$ signifies its decay rate. In the 2D Gray-Scott case, we got 5 trajectories for training, 1 trajectory for validation and 3 trajectories for testing, where $\delta t = 0.5s$. 

\textbf{2D Navier-Stokes Equation.} The Navier-Stokes equations for incompressible flow can be expressed as:
\begin{equation}
\begin{aligned}
  \frac{\partial \mathbf{u}}{\partial t} + (\mathbf{u} \cdot \nabla) \mathbf{u} &= \frac{1}{Re} \nabla^2 \mathbf{u} - \nabla p + \mathbf{f},  \\
  \nabla \cdot \mathbf{u} &= 0.
\end{aligned}
\label{eq:navier_stokes}
\end{equation}

Let $ \mathbf{u} = \{u, v\} \in \mathbb{R}^2 $ denote the fluid velocity vector and $ p \in \mathbb{R} $ the pressure. The Reynolds number ($ Re $) is a dimensionless quantity that characterizes the flow regime, representing the ratio of inertial forces (captured by the advection term $ (\mathbf{u} \cdot \nabla) \mathbf{u} $) to viscous forces (captured by the Laplacian term $ \nabla^2 \mathbf{u} $). 

In this study, training data is generated for $ Re = 1000 $ using a finite volume method (FVM) on a $ 2048^2 $ grid with a simulation time step of $ 2.19 \times 10^{-3} $ s, subject to periodic boundary conditions within the spatial domain $ x \in (0, 2\pi)^2 $. The high-resolution data is then downsampled to a $ 64^2 $ grid to serve as the ground truth. Five trajectories, each consisting of 4800 snapshots, are used for training. For testing, ten trajectories with diverse initial conditions (ICs) are employed for each test case.

\section*{Appendix: Evaluation Metrics}
\label{sectionf}
To assess the performance of our proposed method, we utilize established evaluation metrics: Root Mean Squared Error (RMSE), Mean Absolute Error (MAE), and High Correlation Time (HCT). RMSE quantifies the average magnitude of the difference between predicted and ground truth values, thus serving as a measure of predictive accuracy. MAE quantifies the average absolute difference between predicted and ground truth values, providing a more direct interpretation of the average error. HCT measures the time for which the predicted and true solutions remain highly correlated.

The definitions of these metrics are as follows:

\begin{equation}
\label{metric}
\begin{aligned}
\text{RMSE (Root Mean Square Error):} & \quad \sqrt{\frac{1}{n} \sum_{i=1}^{n} (y_i - \hat{y}_i)^2} \\
\text{MAE (Mean Absolute Error):} & \quad \frac{1}{n} \sum_{i=1}^{n} |y_i - \hat{y}_i|\\
\text{HCT (High Correction Time):}& \sum_{i=1}^{N} \Delta t \cdot 1(\text{PCC}(y_i, \tilde{y}_i) > 0.8)
\end{aligned}
\end{equation}
In the above equations \ref{metric}, $n$ represents the number of trajectories, $y_i$ represents the true value, and $ \hat{y}_i$ represents the predicted value of the model. The $\text{PCC}$ is the Pearson correlation coefficient, which is a statistical metric used to measure the linear correlation between two variables.

\end{document}